# Rumour Detection and Analysis on Twitter

**Yaohou Fan**

University of Melbourne, Melbourne, Australia

Email: yaohouf@student.unimelb.edu.au

Abstract—In recent years people have become increasingly reliant on social media to read news and get information, and some social media users post unsubstantiated information to gain attention. Such information is known as rumours. Nowadays, rumour detection is receiving a growing amount of attention because of the pandemic of the New Coronavirus, which has led to a large number of rumours being spread. In this paper, a Natural Language Processing (NLP) system is built to predict rumours. The best model is applied to the COVID-19 tweets to conduct exploratory data analysis. The contribution of this study is twofold: (1) to compare rumours and facts using state-of-the-art natural language processing models in two dimensions: language structure and propagation route. (2) An analysis of how rumours differ from facts in terms of their lexical use and the emotions they imply. This study shows that linguistic structure is a better feature to distinguish rumours from facts compared to the propagation path. In addition, rumour tweets contain more vocabulary related to politics and negative emotions.

## 1. Introduction

Social media has become more and more important in modern life. On the one hand, social media helps people share information and thoughts; on the other hand, the growth of social media also makes it easier to spread rumours. There are countless rumours on social media like Twitter, and it is necessary to detect them and prevent them from spreading.

There has been a few related works about automatic rumour detection. Recent researches show that RNN based deep learning model can effectively detect misinformation on the Internet [1,2]. The pre-trained BERT [3] model can be a competitive choice to build the rumour detection system. In particular, the BERTWeet [4] trained on 850M sized English tweets may achieve better performance. Proper feature extraction methods can also improve the performance of Bert based classification system [5]. There are also researches that incorporated multiple models such as [6] that combined RNN with CNN to get user features in a time series. Recent studies have shown that the Graph Convolutional Network (GCN) has a remarkable performance in rumour detection, for example, using tree-structured Recursive Neural Networks ( RvNN) for hidden representation of texts and its propagation structure [7], and Bi-Directional GCN (Bi-GCN) specialised for rumour detection on social media to capture propagation and dispersion [8,9].

This research has randomly sampled 1,895 tweets with replies labeled as 'non-rumour' or 'rumour' from three open-source datasets: the PHEME dataset [10], Twitter15 [11], and Twitter16 [12].

## 2. Methodology

*2.1. RNN*

The Recurrent Neural Network (RNN) is proven to be effective for rumour debunking [1,2]. RNN captures the context variations of relevant topics over time, and detects rumours in a quick and accurate way. Because of the nature of the limited context for microblogs, we are interested in the data at an aggregated level. In order to solve the vanishing gradients problem in simple RNN, the Long Short-Term Memory (LSTM) [13] recurrent unit is used in this model, which allows gradients to flow unchanged.

Preprocessing: Each tweet is concatenated with its replies as a string. Then the string is normalised by the Tweet Normalizer, which normalised username, URL and emoji. The model contains the following parts: a text encoder maps input strings into vectors, an embedding layer uses masking to handle the varying sequence lengths, LSTM layers and perception layers. Binary Cross-Entropy is used as the loss function. The Adam optimizer is used to accelerate the gradient descent process.

*2.2. Graph Convolutional Networks*

For a promoting approach, we explored on the graph representation of the text using Graph Convolutional Networks (GCN). GCN learns the features by inspecting neighbouring nodes with various connections and irregular nodes ordering, although a simple GCN model can aggregate the information of relevant posts but loses the sequential orders of rumour propagation trends. We adapted a novel Bi-directional GCN (Bi-GCN) model proposed in [9]. The author introduced the combination of Top-Down GCN (TD-GCN) for rumour propagation and Bottom-Up GCN (BU-GCN) for rumour dispersion (Figure 1). The model also inherits DropEdge and early stopping strategies to avoid overfitting, and enhances the root feature to put more weights on the source tweet. The model achieved an average accuracy of 0.886 and F1 score of 0.930 for rumour class on Twitter 15 dataset in the paper, and outperformed previous methods in terms of all the performance measures.

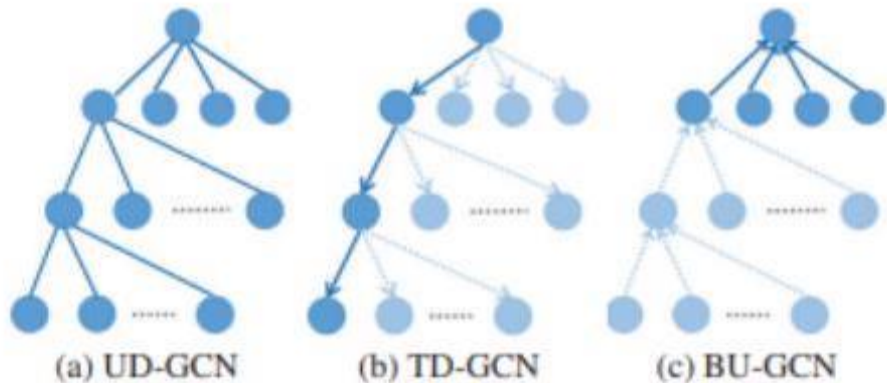

Figure 1: GCN tree structure [9]

Preprocessing: Following a similar structure used in the model, we constructed our tree file with parent index (None for source tweet), child index (from 2 to n for replies in time order), and list of index-count pairs in tab-separation. The index count pairs are calculated using Term FrequencyInverse Data Frequency (TF-IDF), where we construct the frequency metrics using sklearn build in function TfidfVectorizer. We take the top 5,000 words aggregated from the training data and calculate their appearances in each tweet, and store the index and count in a dictionary format. Each tweet with tree structures and labels is saved in a $npz$ format file for further training.

*2.3. BERT*

This research agreed on a hypothesis that the language people use on social platforms such as Twitter differs from other platforms. Nguyen et al. [4] published a pre-trained model on the English corpus named BERTweet. This is a model based on the BERT base [3], trained on 850M sized English tweets with a masked language modeling objective. Their experimental results showed BERTweet outperforms state-of-the-art models on the text classification task. We applied the pytorch framework to implement the BERTweet model.

Architecture: BERTweet has a total parameter size of 110M and contains 12 layers, hidden size of 768 and 12 self-attention heads. Nguyen et al. [4] set the maximum sequence length of BERTweet to 128, so we inherited the same in our task. To provide input to BERTweet, we constructed input IDs and attention mask:

- Input IDs - A sequence of integers which mapping the input tokens to BERTweet tokenizer vocabulary.

- Attention mask - A sequence containing 0 and 1 s, designed to distinguish all padding tokens from input tokens in text of shorter length.

Preprocessing: Taking into account the author's suggestions and combining with our thoughts, we first sorted the text sequence in the order of creation. We also eliminated the sequences missing the original tweets as they lost the purpose of the study. Second, we employ Tweet Tokenizer from the NLTK toolkit to tokenize tweets [14]. Emoticons in the text were transformed into text strings using the emoji package. On the normalization side, we have special tokens for user mentions and URL links, which are @user and HTTPURL respectively. Both tokens are applied to all text.

We also draw inspiration from the Interpersonal Deception Theory [15], which states that readers react differently to deceptive messages than to real content. We thus hypothesise that users on Twitter reply and retweet to rumours using a textual structure or sentiment different from that of non-rumours, e.g. the content of users' replies to rumour might be negative or questioning. Potentially, there are more negatives and question marks in the text than non-rumour. Therefore, we try to separate the original tweets from other derived tweets by dividing each sequence of tweets into original tweets and other tweets, thus forming a sequence of sentence pairs. Segment ids enable the BERT model to learn sentence pairs, and for each token in the text sequence, we specify which sentence it belongs to: original tweets (a series of 0 s) or other tweets (a series of 1 s).

Evaluation: After basic test with $1e^{-5}$ learning rate, $1e^{-2}$ weight decay and $1e^{-7}$ of epsilon value, the accuracy on the development data set reach 0.92, and the Kaggle score is 0.85. It performed slightly worse on rumour dataset (0.80 F1 score) than the non-rumour data (0.93 F1 score). We believe that the main reason for this is the unbalanced class in the data, and a smaller sample of rumour tweets compared to the complexity of BERTweet. However, it still outperformed all other models we have tried. We selected this model as a baseline and conducted parameter tuning for improvements.

## 3. Traditional Machine Learning Models

Apart from the tweet texts, we also experimented on other features. Because of the Twitter regulation and the test data limitation, we only extracted 8 additional features including retweet and like count for the tweet, create year, verified account, followers and following count, tweet count and listed counts for the user. Traditional feature extraction approaches include other handcrafted content-based features such as the number of question marks, different pronouns and emotes, hashtags, mentioned users, and context-based features such as whether the account has a description and URL, whether post on a weekend or weekday, etc. [5]. However, it is very time-consuming to handcraft these features thus out of the scope for this research.

Preprocessing: We inherited the aforementioned Tweet Tokenizer for BERT to produce a numeric representation of the texts. For experiment purpose, max_length is set to 32, 128 and 256 respectively, with attention mask included or excluded. These along with the handcrafted features are fed into data frames for training.

Evaluation: We tested 3 classification methods for this approach, namely random forest, logistic regression and support vector machine. We also introduced class weights and SMOTE upsampling to handle class imbalance. During experiments we found the max_length in encoding and presence of attention mask do not produce much difference on the overall performance, the best result is produced by random forest with SMOTE upsampling and no class weights, the average accuracy is 0.79 with 0.44 F1 score for rumours.

## 4. Experiment Results and Discussion

We decided to use BERT as our final model, and we conducted hyper-parameter tuning to achieve a better accuracy. Table 1 shows the performances of all the models on test set. We repeatedly fine-tune the BERTweet model for a given dataset for 10 epochs. We use AdamW as the optimizer, set the learning rate to $1e^{-4}$ and batchsize 16, and added class weights to resolve the imbalance problem. However, even with same parameters, the model produce different outputs every time. We trained the model multiple times with different combinations of hyper-parameters and used voting to get the most promising results. The final output produces F1 score of 0.96 on the test dataset. Such experimental results prove our previous conjecture that the difference between rumor tweets and truth tweets mainly lies in text features, and although other methods have highlighting performances on some of the measurements, we aim to maximise the accuracy for classifying rumours after all. The result proves the dominant power of pre-trained BERTweet model that is dedicated for rumour identification on Twitter with tweet wording habitations with proper hyper-parameter tuning.

Table 1: Performance of different models

| Method | Class | Acc. | Prec. | Rec. | $F_1$ |
|---|---|---|---|---|---|
| RNN | R | 0.90 | 0.70 | 0.80 | 0.72 |
|  | N |  | 0.93 | 0.94 | 0.93 |
| Bi-GCN | R | 0.90 | 0.79 | 0.77 | 0.78 |
|  | N |  | 0.92 | 0.93 |  |
| Handcraft / RF | R | 0.79 | 0.72 | 0.25 | 0.44 |
|  | N |  | 0.81 | 0.98 | 0.89 |
| Handcraft / SVM | R | 0.69 | 0.30 | 0.24 | 0.27 |
|  | N |  | 0.80 | 0.84 | 0.82 |
| Handcraft / LR | R | 0.77 | 0.68 | 0.11 | 0.19 |
|  | N |  | **0.99** | 0.88 |  |
| BERT (Baseline) | R | 0.92 | 0.82 | 0.80 | 0.80 |
|  | N |  | 0.94 | 0.93 | 0.93 |
| BERT (Fine-tuned) | R | **0.96** | **0.90** | **0.90** | **0.90** |
|  | N |  | 0.96 | 0.97 | **0.96** |

Although the BiGCN model had a decent performance on non-rumours, it did not reach the suggested performance accuracy for rumours as anticipated, and we concluded a few possible reasons as follows: 1). The original model is trained with Twitter15 [11] with a lot more data and includes "reply to the replies", we did not bother to fully construct our tree by tracking those connections among tweets, instead treat all replies as replying to source in time order, this results in our tree to only have one layer and the model is potentially overfitting for such simple setup. 2). We are unaware of how the index-count pairs is calculated in the original setup, a dedicated Twitter word corpus could be more reliable than retrieving word frequencies from training data directly. 3). The rumour and non-rumour classes are imbalanced in our training data. This explains the significant differences in terms of overall accuracy and F1 score for rumours, we could not find suitable resample methods for the raw texts input, converting them into numeric and applying SMOTE does not seem to be effective.

We observed overall deep learning methods outperformed handcrafted features, not to mention the time cost of collecting features. The experiments exhibit the necessity of carrying forward deep learning for NLP.

**5. COVID-19 Rumour Analysis**
In this section, we applied our best-performing BERT model on the COVID-19 related Tweets to conduct analysis on the characters of rumours and non-rumours. We crawled the source tweets of the given ids for simplicity, and combined with COVID-19-related Tweets posted in Australia with the time period from January 2020 to August 2021 retrieved from [16]. A total of 94,222 tweets are used, with 20,068 classified as rumours and 74,154 non-rumours, respectively.

*5.1. Exploratory Data Analysis*
Our first task is to explore the difference between rumours and non-rumours by counting the appearance of different attributes. We examined words, URLs, emojis, hashtags, mentions and stopwords counts in rumour and non-rumours tweets respectively, and plot them into bar charts for analysis (Figure 2). We aim to understand the characteristics of rumours better that can potentially be treated as features for future works.

We identified several patterns from the plots as follows: 1). Word count for non-rumours are distributed evenly, most of the tweets are between 2040 words, whereas the majority of rumours have less than 20 words. More words also indicate more stopwords are used. 2). URL is not preferred by both rumours and non-rumours. About half of the tweets did not include any URLs, and on average, nonrumour tweets prefer to include more URLs than rumours. 3). There are no significant differences between emojis, mentions, and hashtags included in rumours and non-rumours. The mojority of tweets included 0-5 mentions and hashtags and 0-1 emojis. The findings are not surprising and validated our hypothesis, the wording and phrasing are more formal and well-examined for non-rumours compared to unverified daily Twitter users who are inclined to share their emotions with casual languages and internet buzzwords.

*5.2. Word Cloud for Topic Trends over Time*
We split the data into one-month intervals and use the word cloud package to find heated discussion keywords as shown in figure 3. We categorised the tweets and extracted some keywords as themes, expecting to explore the changes. Note that we removed some basic keywords such as 'covid', 'corona virus' for better comparison. A large number of tweets were collected from Australia, so the results of the experiment are regional in nature.

In early 2020, the themes of deceptive and non-deceptive tweets overlapped, both revolving around 'Wuhan' and 'China', indicating the origin of the pandemic. In rumour tweets, we noticed some special themes such as 'create', 'lab' and 'beer'. Over the next six months, 'American', 'Trump' and their synonyms gradually replaced

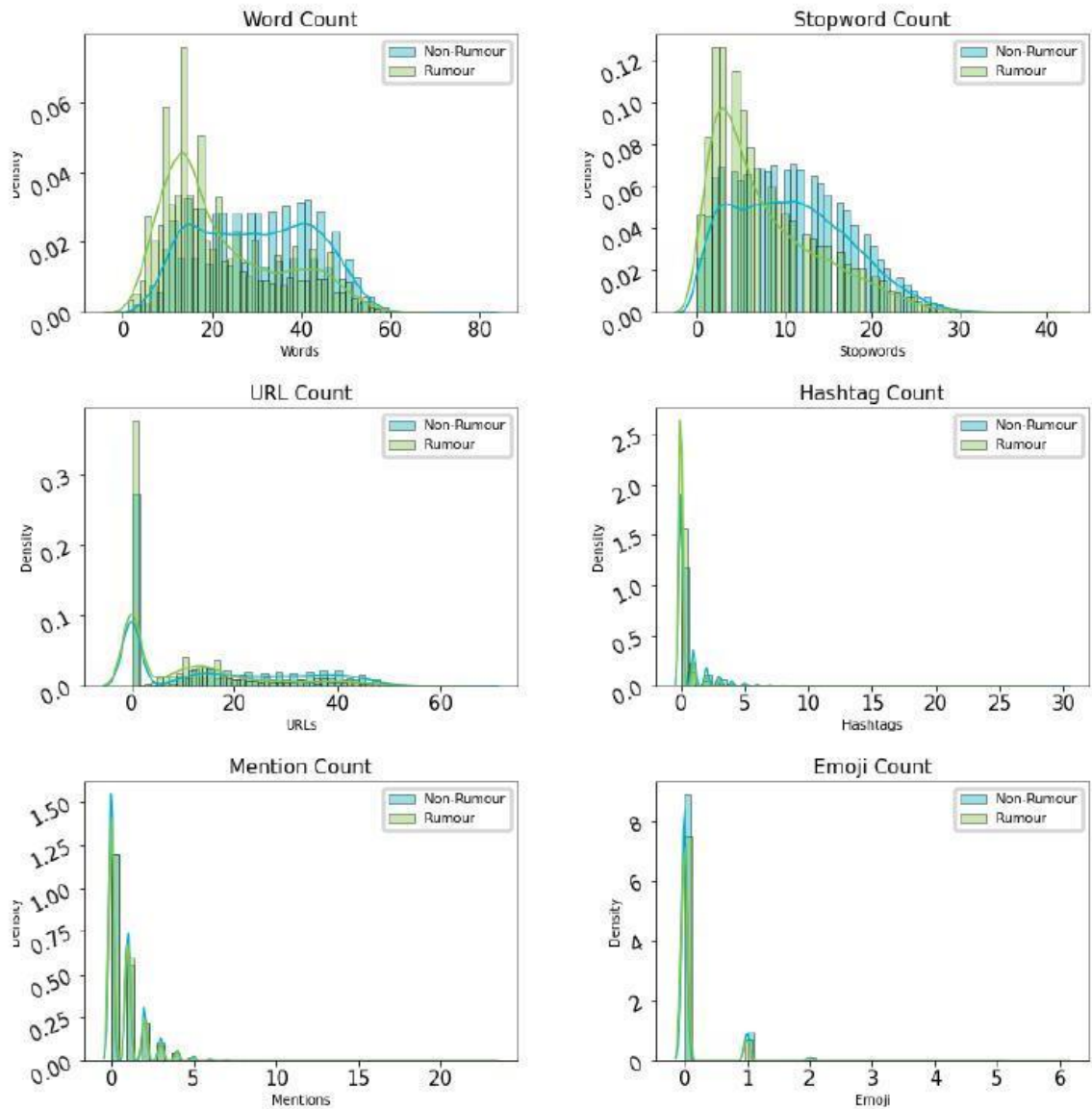

Figure 2: Exploratory analysis rumours & non-rumours

'Wuhan' and 'China' and became the new topic of conversation. Although neutral words such as "new" and "pandemic" are top words in both categories, they appear significantly more frequently in non-rumours than in rumours. The opposite trend was observed for words with exaggerated meanings such as "death", "many" and "million". We also spotlighted 'Trump', which first appeared in rumours tweets in February and became the most popular word during the following month. In the year since, 'trump' has been one of the top words in rumours. From July, words like 'lockdown', 'aged care', and 'crises' started to appear, the pandemic continued to evolve and began to draw more people's attention, with anxiety and dissatisfaction.

In 2021, a significant number of Twitter users continued to post around the "new case". What caught our attention was the fact that "vaccine" became one of the most frequently discussed topics among Twitter users, both in rumours and non-rumours. Some breaking news about COVID-19 was also one of the top words of the month, such as the name "Craig Kelly" for an Australian MP appearing in February 2021.

In summary, the experimental analysis shows the highly overlapping nature of rumour and non-rumour themes. In the overlap interval, neutral terms appear more frequently in true and reliable tweets than in rumours, and conversely, negative terms and specific terms are more likely to appear in rumours than in truths. In addition, breaking news is a potential topic of rumours.

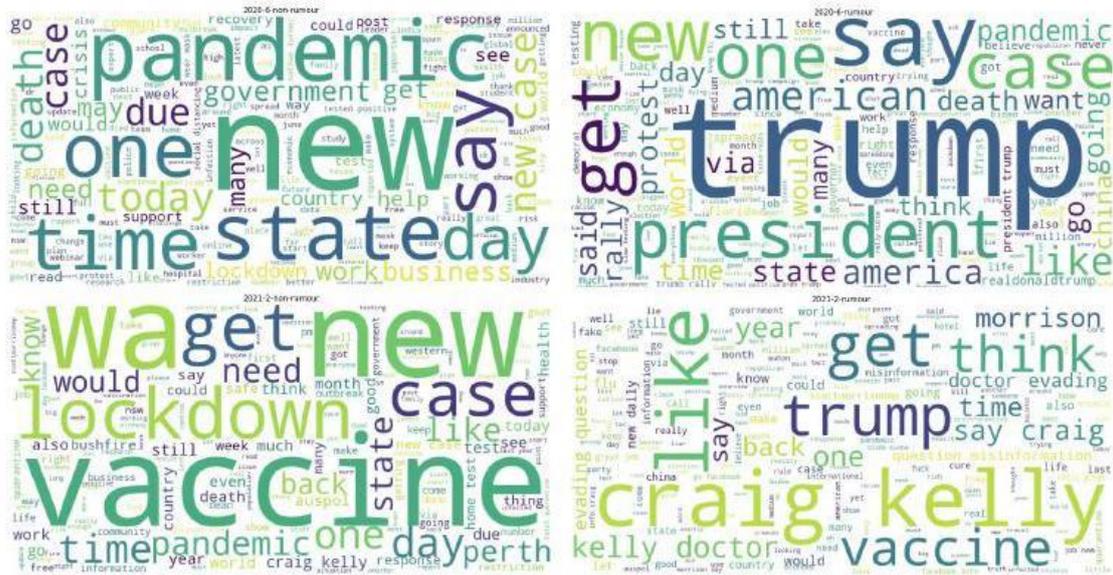

Figure 3: Word cloud highlights non-rumour(L) & rumour (R)

*5.3. Emotion and Sentiment Analysis*

Finally, we looked at the emotions and sentiment levels of tweets. Such analysis has always been an important territory for NLP studies, and we are keen for interesting findings. We used text2emotion package that convert the texts into numeric representations of different emotions including happy, angry, surprise, sad and fear, and we label each tweet with the emotion with the highest score. Similarly, we used Sentiment Intensity Analyzer in NLTK package to conduct sentiment analysis (Figure 4).

The distribution of emotions and sentiments is moderately similar for rumours and non-rumours tweets. For both rumours and non-rumours, we noticed fear and surprise took a large proportion, especially for non-rumours which fear has taken up 36%. More negative tweets are found in rumours than non-rumours, which all illustrated people's worry and anxiety for this unknown pandemic. Surprisingly more neutral tweets are found in rumours than non-rumours. In terms of the average score perspective, fear is always the one with the highest score during the pandemic, alerting us the danger of pandemic and not to relax our vigilance. The average score for non-rumours is substantially stable. For rumours, major fluctuations appear in fear, sad and surprise. More optimism is given during the end of 2020, people are presumably celebrating Christmas and New Year, leaving behind the sadness from COVID for a moment.

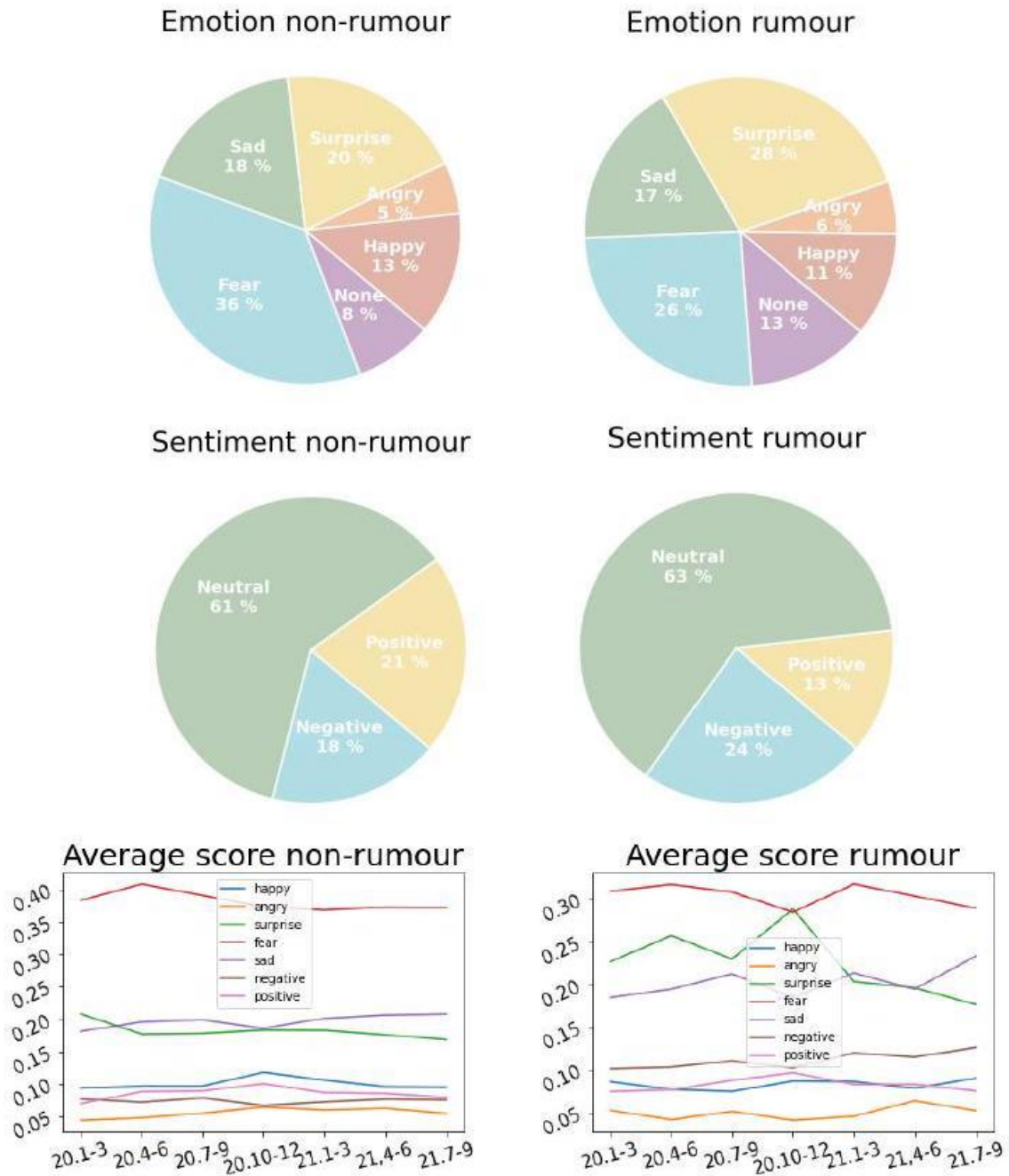

Figure 4: Emotion and sentiment distribution & Average score over time

**6. Conclusion**

In this paper, we explored 4 machine learning NLP models to conduct rumour and non-rumour classifications on Twitter. Our best model uses finetuned pretrained BERTweet and scores 0.96 on test set. The experimental results demonstrate that the GCN model does not perform as well as BERTweet because it captures the path of rumour propagation, while the latter captures the semantics in the tweets. This suggests that the semantics of tweets is the most effective information to distinguish rumours from facts. After this, we applied our model to COVID-19-related tweets and conducted exploratory analysis on the characteristics of rumours and non-rumours, as well as the trending topics over time and emotions and sentiment analysis. We found that between 2020 and 2021 the themes of rumours and facts are highly overlapping, with neutral terms appearing more often in reliable tweets, while negative and politically relevant terms appear mostly in rumours. In addition, breaking news is a potential topic for rumours. The content and themes of rumour propagation change over time. However, a limitation of the model used in this study is that it cannot detect rumours online in real-time. Detecting rumours accurately and in real-time for the vast amount of information available on social platforms may become a focus of future research.